\documentclass[letterpaper]{article} 
\usepackage{aaai25}  
\usepackage{times}  
\usepackage{helvet}  
\usepackage{courier}  
\usepackage[hyphens]{url}  
\usepackage{graphicx} 
\usepackage{natbib}  
\usepackage{caption} 
\usepackage{algorithm}
\usepackage{algorithmic}

\usepackage{booktabs} 
\usepackage{multirow}
\usepackage{amsfonts}

\usepackage{newfloat}
\usepackage{listings}
\DeclareCaptionStyle{ruled}{labelfont=normalfont,labelsep=colon,strut=off} 
\floatstyle{ruled}
\newfloat{listing}{tb}{lst}{}
\floatname{listing}{Listing}
\title{HieraFashDiff: Hierarchical Fashion Design with Multi-stage Diffusion Models}
\author{
    Zhifeng Xie\textsuperscript{\rm 1,\rm 4},
    Hao Li\textsuperscript{\rm 1},
    Huiming Ding\textsuperscript{\rm 1},
    Mengtian Li\textsuperscript{\rm 1,\rm 4},
    Xinhan Di\textsuperscript{\rm 3},
    Ying Cao\textsuperscript{\rm 2}\thanks{Corresponding author.}
}
\affiliations{
    \textsuperscript{\rm 1}Department of Film and Television Engineering, Shanghai University\\
    \textsuperscript{\rm 2}School of Information Science and Technology, ShanghaiTech University\\
    \textsuperscript{\rm 3}AI Lab, Giant Network\\
    \textsuperscript{\rm 4}Shanghai Engineering Research Center of Motion Picture Special Effects\\

    \{zhifeng\_xie, dinghuiming, mtli\}@shu.edu.cn, hao.li.shu@outlook.com, dixinhan@ztgame.com, caoying59@gmail.com

}

\usepackage{bibentry}

\begin{document}

\maketitle

\begin{abstract}
Fashion design is a challenging and complex process.
Recent works on fashion generation and editing are all agnostic of the actual fashion design process, which limits their usage in practice.
In this paper, we propose a novel hierarchical diffusion-based framework tailored for fashion design, coined as \textit{HieraFashDiff}. 
Our model is designed to mimic the practical fashion design workflow, by unraveling the denosing process into two successive stages: 1) an ideation stage that generates design proposals given high-level concepts and 2) an iteration stage that continuously refines the proposals using low-level attributes. 
Our model supports fashion design generation and fine-grained local editing in a single framework. 
To train our model, we contribute a new dataset of full-body fashion images annotated with hierarchical text descriptions. 
Extensive evaluations show that, as compared to prior approaches, our method can generate fashion designs and edited results with higher fidelity and better prompt adherence, showing its promising potential to augment the practical fashion design workflow. 
Code and Dataset are available at \url{https://github.com/haoli-zbdbc/hierafashdiff}.
\end{abstract}

%

\section{Introduction}

\begin{figure*}[!t]
  \includegraphics[width=1.0\textwidth]{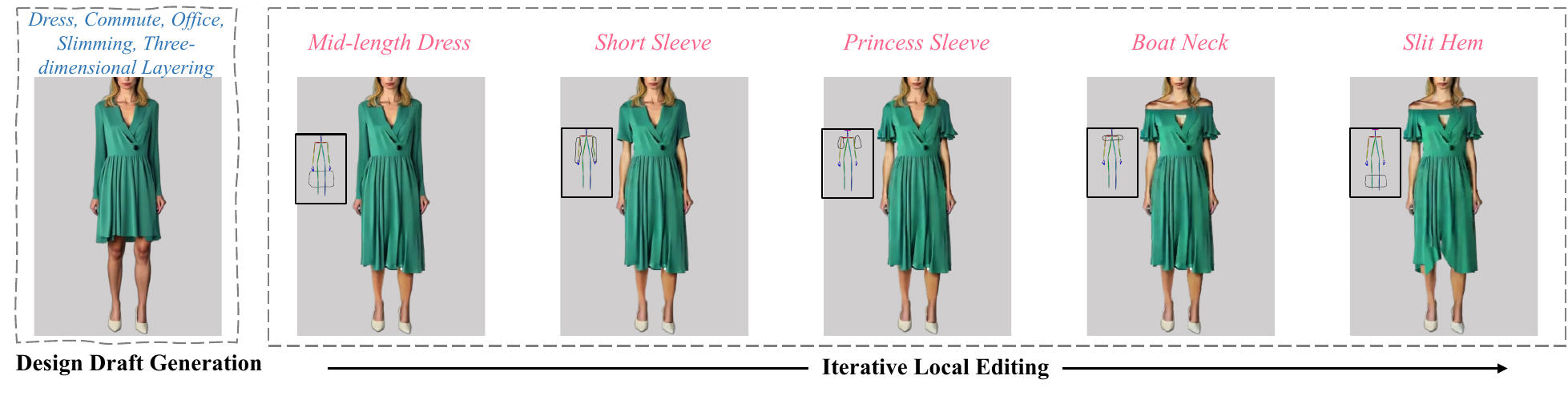}
  \centering
  \caption{The proposed HieraFashDiff is capable of generating fashion design drafts from just abstract concepts (blue text), and allowing for local editing on the generated draft iteratively through a few apparel attribute descriptions (red text). 
  Thus, our method can be used to facilitate typical fashion design workflow by enabling efficient ideation and rapid iteration.}
  \label{fig:teaser}
\end{figure*}

Fashion design is an important and challenging activity, which requires navigating through a huge joint space of shape, color, material, pattern and layout in order to find solutions that satisfy aesthetic and functional requirements, and perhaps client-specific constraints. 
The typical workflow of fashion designers consists of two essential stages: \textit{ideation} and \textit{iteration}.
In particular, fashion designers begin their creative process by coming up with some abstract design concepts in terms of theme, style and personality, and then translating these concepts into concrete design ideas (i.e, design drafts) through brainstorming or turning to reference examples for inspiration.
Subsequently, they iterate on the initial idea by making small changes to the draft to finalize the design.

Recently, there is a growing interest in facilitating the fashion design process by building generative models that can synthesize realistic fashion images from user-specified hints and constraints~\cite{chen2020tailorgan,dai2021edit,cao2023image}. 
However, these approaches is that they do not consider (and thus fail to fit into) the practical fashion design pipeline, making them difficult to be directly adopted in practical scenarios.

Building models that can find wide adoption in the real fashion design process is non-trivial. 
First, a desirable model should learn the task of fashion design generation and editing jointly to separately tackle the ideation and iteration stages, so that it has capability to support the entire pipeline. 
This is in contrast to prior works that especially address either generation~\cite{zhu2017your,jiang2022text2human,zhang2022armani,sun2023sgdiff} or editing~\cite{ak2019attribute,kwon2022tailor,pernuvs2023fice,baldrati2023multimodal,wang2024texfit}. 
Second, both generation and editing components should be conditioned on intuitive inputs at high and low levels, respectively, empowering users to express their design thoughts easily while facilitating efficient ideation and rapid iteration.

To address the aforementioned challenges, we propose \textit{HieraFashDiff}, a novel framework specialized in facilitating fashion design, which instantiates a conditional diffusion model that learns to generate fashion images from given fashion-specific text prompts. 
Our key insight is that the typical fashion design workflow is reminiscent of the reverse process of diffusion models for image generation, which first denoises a purse noise into a coarse image (design draft) in the early stage, and then iteratively refine the coarse image to add more fine details to produce a realistic image (final design) over the remaining steps. 
Inspired by this, we propose to factor the reverse process of our diffusion model into two successive stages: an ideation stage spanning the earlier denoising steps and an iteration stage spanning the later denoising steps. 
Our model injects text descriptions at different levels into the two stages \textemdash ~the ideation stage is conditioned on high-level design concepts to produce noisy design drafts while the iteration stage is guided by progressively added low-level apparel attributes to refine the drafts towards complete designs.
In this way, our framework naturally supports the concept-guided generation of fashion design proposals and the semantic editing of local fashion components in a unified framework to aid in the whole fashion design pipeline, as shown in Fig \ref{fig:teaser}.

To train our model, we curate a new fashion dataset, coined as \textit{HieraFashion}.
Our dataset consists of more than 5k full-body apparel images, each of which is annotated with a hierarchical text caption that is concise yet informative enough to capture the essence of fashion design. 
We evaluate our model on our newly collected dataset, demonstrating its state-of-the-art performance in terms of generation quality and prompt coherence, as compared to existing methods.

In summary, our contributions are as follows.
\begin{itemize}
    \item {We propose a fashion generation and editing framework that, for the first time, mimics the \textit{whole} fashion design process explicitly, which can support efficient ideation and rapid iteration.}
    \item {We propose a novel hierarchical text-to-fashion diffusion model that decomposes the generation process into multiple stages conditioned on input prompts of different levels, which enables coarse-grained fashion draft generation and fine-grained modifications \textit{jointly}.} 
    \item {We curate a new dataset comprising full-body fashion images captioned with high-level design concepts and low-level local attributes.}
\end{itemize}

\section{Related Work}

{\bf Text-Guided Fashion Image Generation.}
Text-to-image generation is a crucial and complex task that seeks to generate realistic images based on natural language descriptions.
In the fashion domain, only a few works~\cite{zhu2017your,zhang2022armani,sun2023sgdiff} attempt to generate fashion-related images (e.g, for apparel, accessories and fashion models) from textual description.
Early approach~\cite{zhu2017your} to the text-guided fashions synthesis relied on Generative Adversarial Networks(GANs) presented a two-stage stylized image generation solution that generates realistic fashion images, conditioned on textual descriptions and semantic layouts.
Zhang et al~\cite{zhang2022armani} proposed the ARMANI framework for fashion synthesis focused on generating local details. 
Recent advances in diffusion models~\cite{nichol2021glide,rombach2022high,ramesh2022hierarchical,saharia2022photorealistic} lead to more realistic generation. 
Sun et al~\cite{sun2023sgdiff} developed and applied the skip cross-attention module to integrate image and text modalities. 
Our method adopts a multi-stage framework to closely mimic the practical graphic design workflow, enabling it to support fashion generation and editing simultaneously.
Moreover, our method handles hierarchical text descriptions with explicit disentanglement of high-level concepts and low-level attributes, instead of captions that either encompass only low-level details or mix up high-level and low-level prompts together, to better facilitate ideation and iteration in fashion design.

\noindent {\bf Fashion Image Editing.}
Generative Adversarial Networks (GANs) have emerged as a cornerstone technology extensively applied in fashion-related image editing tasks~\cite{ak2019attribute,kwon2022tailor}. FICE~\cite{pernuvs2023fice} addressed text-conditional image editing with optimization-based GAN inversion guided by the CLIP model.
Some recent efforts started to approach fashion image editing diffusion models.
MGD ~\cite{baldrati2023multimodal} proposed a latent diffusion model to edit fashion images conditioned on multimodal inputs including text, pose and sketch.
Zhang et al~\cite{zhang2023diffcloth} introduced a diffusion model incorporating structural semantic consensus guidance, utilizing a language structure parser to extract attribute words, thereby achieving fine-grained semantic alignment.
TexFit ~\cite{wang2024texfit} predicted the editing area in an image based on the input text and used the predicted region to condition a diffusion model for local editing.
Our method can perform local editing iteratively on full-body fashion images from apparel attribute descriptions, producing a sequence of high-quality, continuously evolving designs, which has not yet been demonstrated in the existing works. 
In addition, our method complements its editing functionality with a capability to generate full-body fashion designs from just high-level concepts, which is not available in the previous approaches.

\section{Method}

\begin{figure*}[!t]
  \centering
   \includegraphics[width=0.85\linewidth]{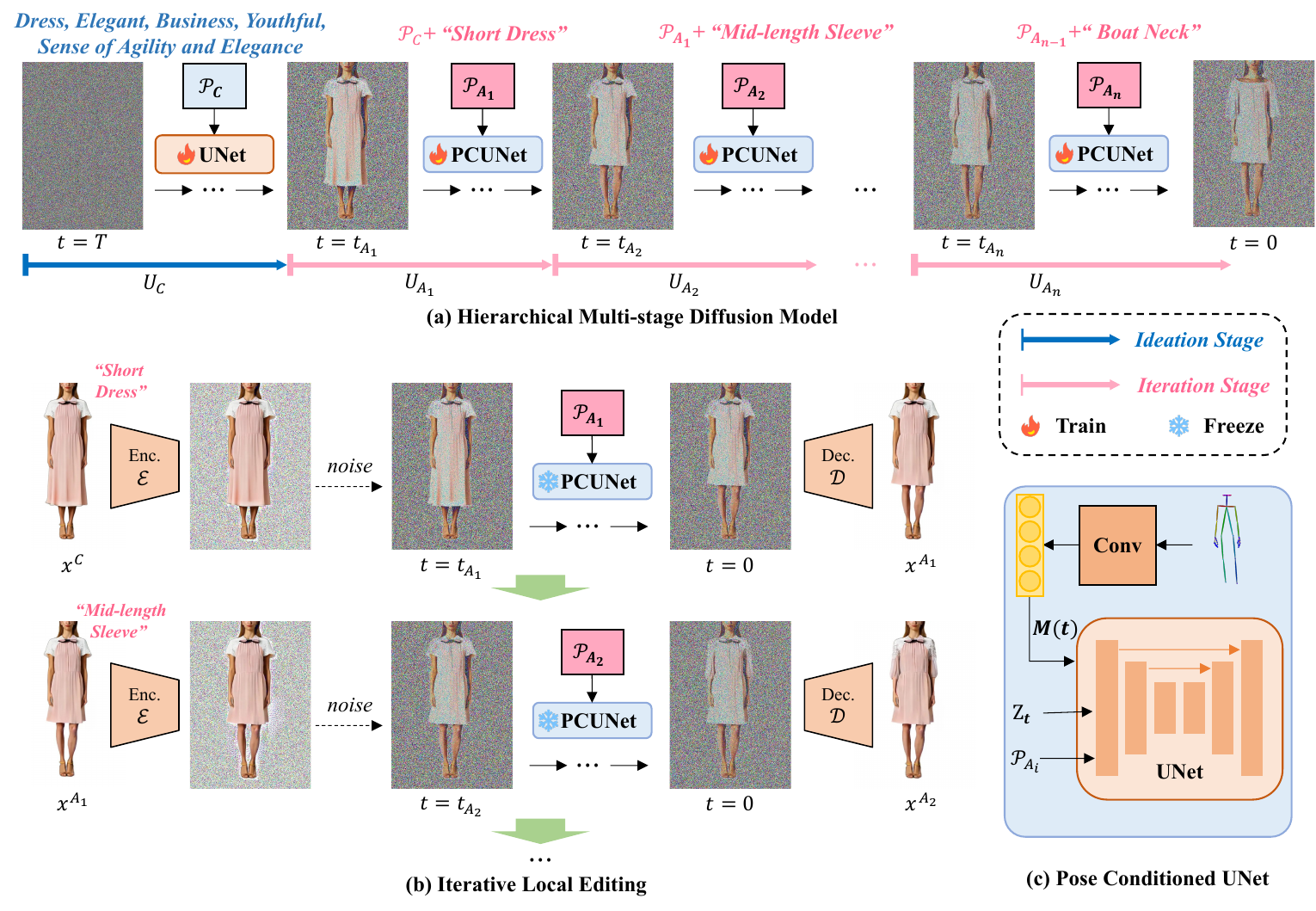}
   \caption{Overview of our method. 
   (a) The denoising process of our model is decomposed into an ideation stage and an iteration stage, which are conditioned on high-level concepts and low-level attributes, respectively.
   (b) our editing method starts from the generated design daft $x^C$ and produces a sequence of edited results $(x^{A_1}, x^{A_2}, \dots)$ given text prompts for different attributes $(A_1, A_2, \dots)$.
   (c) our UNet-based denoising network is conditioned on additional pose information.}
   \label{fig:method}
\end{figure*}

\subsection{Hierarchical Multi-stage Diffusion Model}
We build our model on the pre-trained Stable Diffusion (SD)~\cite{rombach2022high} and fine-tune the SD on a fashion dataset~\cite{baldrati2023multimodal} and, therefore, our model operates in the latent space instead of pixel space. 
The key idea underlying our model is to abstract the common fashion design workflow as the denoising process of our model. 
As illustrated in Fig \ref{fig:method}, the denoising process (i.e, the generation process) is decoupled into two sequential, successive stages: an ideation stage and an iteration stage.

\noindent\textbf{Denoising Process Decomposition.} 
Let $U$ denote the full timesteps of the denoising process. 
The ideation and iteration stages span the earlier steps $U_C$ and the later steps $U_A$, respectively, so that $U=U_C \cup U_A$. 
The ideation stage is responsible for generating a design daft from high-level, vague concepts $\mathcal{P}_C$, and the iteration stage aims to iteratively refine the generated draft based on low-level attribute descriptions $\mathcal{P}_A$, by adding more and more fine-grained, local design components and details. 
Such decomposition formulation allows our model to naturally support automatic design generation and semantic design editing in an unified manner.

To enable more localized control in the editing scenario, we further decompose the iteration stage into a sequence of sub-stages, one for each apparel attribute \textemdash ~$U_A = U_{A_1} \cup U_{A_2} \cup \dots \cup U_{A_n}$, where $U_{A_i}$ represents the time interval dedicated to $i$-th attribute and $n$ is the number of attributes being considered. 
In this way, it is possible for users to iterate on the generated design by solely changing one attribute at a time. 
To arrange the apparel attributes sequentially through the denoising process, we need to determine their ordering.
To this end, we leverage the inherent property of the diffusion model to maximize  editability for each attribute.
In particular, due to the noise variance schedule~\cite{ho2020denoising}, the amount of change to a generated image declines over denoising steps~\cite{cao2023masactrl,yu2023freedom}. 
Therefore, we choose to order the attributes by the size of their affected areas in fashion images. 
For example, modifying dress length will cause a larger proportion of an images to be changed than modifying neckline type, and thus we rank dress length before neckline type. 
This will ensure that our model has sufficient ability to make desired edits to faithfully reflect different attributes.
In our implementation, we consider 5 common apparel attributes, and order them as: ``clothing length" $>$ ``sleeve length" $>$ ``sleeve type" $>$ ``collar type" $>$ ``hem type".

\noindent\textbf{Prompt Schedule.} 
During the denoising process, the ideation stage is conditioned on the abstract design concepts $\mathcal{P}_C$, and the iteration stage are conditioned on a sequence of text prompts $\mathcal{P}_A = (\mathcal{P}_{A_1}, \dots, \mathcal{P}_{A_n})$, where each sub-stage $i$ only generates local component given $\mathcal{P}_{A_i}$. 
We construct the iteration sub-stage prompts by starting with $\mathcal{P}_{C}$ and adding one attribute description per sub-stage over time. 
Formally, the prompt for $i$-th iteration stage is defined as:
\begin{equation}
    \mathcal{P}_{A_i}=\mathcal{P}_C+\sum_{k=1}^iA_k,
  \label{eq:eq4}
\end{equation}
where the addition operator is overloaded to denote the concatenation of two prompt strings. 
This means that each iteration sub-stage needs to consider not only its own attribute but also the given concepts and all the previous attributes, thereby helping better preserve what is already generated.

\noindent\textbf{Conditioning on Pose Map.} 
During the iteration stage, besides text prompts, we also input a 2D pose map, encoding 18 body joint positions, into our denoising network to further improve generation quality. 
The pose map is obtained by running a 2D pose detector~\cite{cao2017realtime} on the real image in training and the generated image by the ideation stage in testing.
Providing joint position information to the model can help it better localize the regions to change for a given attribute. 
Furthermore, we find this additional input is beneficial to the preservation of the human pose during editing, as observed in ~\cite{baldrati2023multimodal}.

\noindent\textbf{Training.} 
For training, we optimize the following objective:
\begin{equation}
  \mathbb{E}_{z_0, t, \epsilon}\left[\left\|\epsilon-\epsilon_{\theta}\left({z}_{t}, t, \mathcal{C}(t), \mathcal{M}(t)\right)\right\|_{2}^{2}\right],
\end{equation}
where $\mathcal{C}(t)$ is prompt schedule function:
\begin{equation}
    \mathcal{C}(t) = 
    \left\{\begin{array}{lr}
     \mathcal{P}_{C}, & t \in U_C \vspace{5pt}\\ 
     \mathcal{P}_{C} + \sum_{k=1}^tA_k, & t \in U_{A_t}
    \end{array}\right.
    \label{eq:eq5}.
\end{equation}
$\mathcal{M}(t)$ is a pose schedule function that returns the 2D pose map of the training image. if $t$ falls within the iteration phase and empty if $t$ is within the ideation phase.

\subsection{Design Synthesis and Editing}   

\noindent{\bf Design Draft Generation.}
Given a high-level concept prompt $\mathcal{P}_C$, we aim to generate a design draft. 
To do this, we sample a base noise $z^C_T$ and feed it into our denoising process to obtain $z^C_0$, which is then decoded by the SD decoder $\mathcal{D}$ to generate a design draft $x^C = \mathcal{D}(z^C_0)$. 
Note that our sampling is run through all the timesteps, i.e, from $t=T$ to $t=0$, instead of just through the ideation stage, in order to generate a \textit{clear} design draft. 
This is possible because our denoising network sees the high-level concept description for all the timesetps during training according to our prompt schedule function in Eq. \ref{eq:eq5}.

\noindent{\bf Iterative Local Editing.}
Given the generated draft $x^C$, we aim to edit it iteratively through a sequence of intuitive apparel attribute descriptions $(\mathcal{P}_{A_1}, \dots, \mathcal{P}_{A_n})$. 
As shown in Fig \ref{fig:method}, to perform editing on the first attribute $A_1$, we take the generated draft $x^C$ as input and diffuse its latent obtained using the pre-trained image encoder to a noisy latent at the starting timestep, $t_{A_1}$, of the first attribute sub-stage using the forward process margin distribution $q(z_t|z_0)$. 
Then, we run the rest sampling steps to $t=0$ from the noisy latent conditioned on $\mathcal{P}_{A_1}$ to produce the edited result $x^{A_1}$. 
For each subsequent attribute $A_k$, we execute editing in a similar way except that the input is the edited result of the previous attribute and the sampling is run from the starting timestep of attribute $A_k$ with $\mathcal{P}_{A_k}$ as input prompt.

Typically, modifying an attribute, which only refers to a local \textit{target} region in the fashion image, should ideally lead to localized changes. 
However, We find that the above method may cause undesirable non-local changes.
To mitigate this issue, we observe that there exists a strong correlation between apparel attributes and human body parts \textemdash ~e.g, changing sleeve type will primarily influence the region on and near the arm. 
This motivates us to leverage human body part masks associated with the corresponding attributes to enforce the preservation of non-target regions. 
To generate the body part masks, we apply the 2D pose detection method~\cite{cao2017realtime} to the input image to estimate joint locations.
For a body part associated with attribute $a$, we construct a bounding box enclosing the relevant joint positions, and use the SAM~\cite{kirillov2023segment} to segment the clothing region within the bounding box to form a binary mask, which labels the target and non-target regions with 1 and 0, respectively. 
Note that when increasing the length of an apparel component (e.g, from short dress to long dress), we directly use the entire region within the bounding box as the mask to cover the non-clothing region that needs to be modified (e.g, leg).
Then, similar to the blended latent diffusion~\cite{avrahami2023blended}, we modify each denoising step within the period of attribute $a$ using its mask $\mathbf{m}_a$ (downsampled to the spatial resolution of the latent):
$\tilde{\mathbf{z}}_{t-1}=\mathbf{m}_a\odot\mathbf{z}_{t-1}+(\mathbf{1}-\mathbf{m}_a)\odot \mathbf{z}^i_{t-1}$, 
where $\mathbf{z}_{t-1}$ is sampled from the learned condition distribution $p_\theta(\mathbf{z}_{t-1}|\mathbf{z}_{t)}$, $\odot$ denotes element-wise multiplication and $\mathbf{1}$ is an all-ones image. 
$\mathbf{z}^i_{t-1}$ is a noisy latent obtained by encoding the input image via $\mathcal{E}$ into a latent and diffusing it to timestep $t-1$. 
Intuitively, this keeps the non-target region unchanged, by replacing the values of $\mathbf{z}_{t-1}$ in the non-target region with their counterparts from the corrupted input image encoding.

\section{Hierarchical Fashion Dataset}

To train our model, a fashion dataset captioned by both high-level concepts and low-level attributes is needed.
Existing fashion datasets~\cite{yang2020fashion,morelli2022dress,jiang2022text2human,baldrati2023multimodal} is not sufficient to serve our purpose. 
Their texts captions either lack a complete capture of fashion design concepts or mix up the descriptions of different levels together without explicit separation. 
Therefore, we curate a dataset, \textit{HieraFashion}, with 5200 full-body fashion images of high resolution (768×1024 pixels), spanning 8 common apparel categories including dress, coat, sweater, blouse, jumpsuit, pant, shirt and skirt.

The images in our dataset are meticulously selected from FACAD~\cite{yang2020fashion}, DeepFashion-MM~\cite{jiang2022text2human} and Dress Code-MM~\cite{baldrati2023multimodal} so that they have diverse design elements and rich variation in design pattern. 
To ensure the visual quality of the images and consistency across them, we cropped out the area above the eyes of the model in each image, and filter out the images with complex backgrounds, non-frontal body orientations, challenging poses, and half-body shots.
One notable feature of our dataset is its hierarchical text descriptions \textemdash ~each image is annotated with a text description which comprises two parts: high-level design concepts and low-level apparel attributes. The textual annotations were performed by recruited \textit{professional} fashion designers.

Overall, our dataset has several distinctive characteristics: 1) hierarchical text descriptions; 2) full-body fashion images with clear background; 3) high-resolution images. 
Please refer to the supplemental materials for more details.

\begin{figure*}[t]
  \centering
   \includegraphics[width=0.85\linewidth]{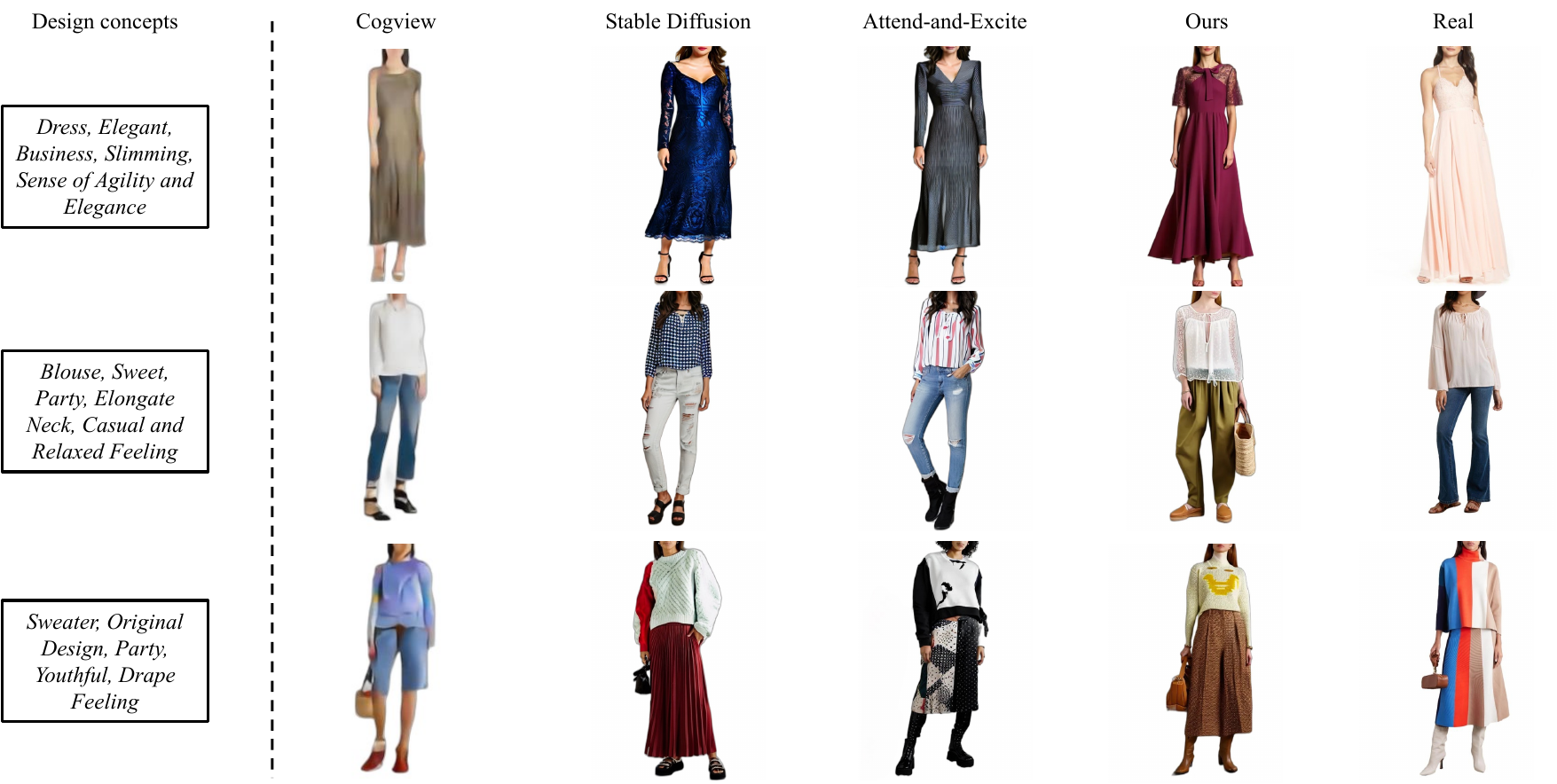}
   \caption{Qualitative results of different methods for fashion draft generation from high-level design concepts.}
   \label{fig:generation}
\end{figure*}

\section{Experiments}

{\bf Datasets. }
We conduct experiments on the \textit{HieraFashion} dataset.
We divide the \textit{HieraFashion} dataset into 4,000 training examples and 1,200 testing examples.
Further, we use the Dress Code Multimodal dataset, with a total of 26, 400 image-text pairs to fine-tune the models.

\noindent{\bf Implementation Details. }
We fine-tune the stable diffusion model on the Dress Code Multimodal and \textit{HieraFashion} datasets. 
We use 1000 timesteps for the reverse process, allocating the interval[1000, 900] to the ideation stage and setting the intervals for the 5 attributes in the iteration stage as [900, 800], [800, 710], [710, 630], [630, 560], [560, 0]. 
Note that we put most of the iteration sub-stages into the first half of the denoising process since the model has very restricted flexibility to change the generated image in the late denoising stage~\cite{cao2023masactrl,yu2023freedom}.
During training, we resize all the images to 512×704.
We train our model for 117,299 steps on a single NVIDIA A6000 GPU on our \textit{HieraFashion} dataset, employing a batch size of 4, a learning rate of 1e-5, and a linear warm-up for initial 500 iterations, with the AdamW~\cite{LoshchilovH19} optimizer.
For image generation, we employ DDIM~\cite{song2020denoising} with 100 steps as noise scheduler and use classifer-free guidance~\cite{ho2022classifier}.

\begin{table}
    \centering
    \scalebox{0.9}{
    \begin{tabular}{ccccc}
        \toprule 
        \textbf{Method} &  & \textbf{FID} $\downarrow$  & \textbf{Coverage} $\uparrow$ & \textbf{CLIP-S} $\uparrow$ \\
        \midrule
        Cogview &  & 23.62 & 0.35 & 25.87  \\
        SD-finetune &  & 18.45 & 0.48 & 27.92 \\
        Attend-and-Excite &  & 15.33 & 0.63 & 28.46  \\
        \midrule
         Ours &  & \textbf{10.27} & \textbf{0.76} & \textbf{30.61} \\
         \bottomrule
    \end{tabular}
    }
    \caption{Quantitative evaluation of different methods for fashion draft generation. The best results are in bold.}
    \label{tab:generation}
\end{table}

\noindent{\bf Compared methods. }
For design draft generation,  we compare our method with state-of-the-art methods including Cogview~\cite{ding2021cogview}, Stable Diffusion~\cite{rombach2022high}, and Attend-and-Excite~\cite{chefer2023attend}. 
To ensure a fair comparison, all models are retrained using the Dress Code Multimodal~\cite{baldrati2023multimodal} and \textit{HieraFashion} datasets.
For local editing, we consider general-purpose image inpainting methods, SD-Inpaint\footnote{https://huggingface.co/runwayml/stable-diffusion-inpainting} and BrushNet~\cite{ju2024brushnet}, as baselines.
We also compare with latest text-guided fashion image editing methods: FICE~\cite{pernuvs2023fice}, MGD~\cite{baldrati2023multimodal}, TexFit~\cite{wang2024texfit}.
Both FICE and TexFit are retrained on the \textit{HieraFashion} training set; MGD is not trained on our dataset since its training code is not accessible. 
For fair comparison, MGD, BrushNet and TexFit use the same body part masks as our method.
We acknowledge that there are more generation and editing methods~\cite{zhang2022armani,li2022m6,sun2023sgdiff,zhang2023diffcloth} in the fashion domain.
However, comparison with them is not feasible since their training code is not publicly available.

\begin{table*}
    \centering
    \scalebox{0.9}{
    \begin{tabular}{cccccccccccccccc}
    \hline 
     \multirow{2}{*}{\textbf{Method}}  & & \multicolumn{2}{c}{$\mathbf{A_1}$} & & \multicolumn{2}{c}{$\mathbf{A_2}$} & & \multicolumn{2}{c}{$\mathbf{A_3}$} & & \multicolumn{2}{c}{$\mathbf{A_4}$} & & \multicolumn{2}{c}{$\mathbf{A_5}$} \\
     & & \textbf{FID} $\downarrow$  & \textbf{CLIP-S} $\uparrow$ & & \textbf{FID} $\downarrow$  & \textbf{CLIP-S} $\uparrow$ & & \textbf{FID} $\downarrow$  & \textbf{CLIP-S} $\uparrow$ & & \textbf{FID} $\downarrow$  & \textbf{CLIP-S} $\uparrow$ & & \textbf{FID} $\downarrow$  & \textbf{CLIP-S} $\uparrow$ \\
     \midrule
     FICE & &22.72 &27.94 & &23.15 &28.07 & &23.69 &28.13 & & 24.28 & 28.36 & & 25.53 & 28.47  \\
     SD-Inpaint  & &13.56 & 28.13 & & 14.31 & 28.24 & & 15.57 & 28.32 & & 16.76 & 28.54 & & 17.28 & 28.66 \\
     TexFit & &13.98 &28.01 & &14.82 &28.17 & &15.73 &28.15 & & 16.50 & 28.42 & & 17.11 & 28.64  \\
     TexFit-M & &13.27 &28.59 & &13.83 &28.76 & &14.41 &28.92 & & 15.16 & 30.38 & & 15.72 & 30.55  \\
     BrushNet & &12.92 &29.85 & &13.46 &30.10 & &13.98 &30.54 & & 14.63 & 31.12 & & 15.06 & 31.31  \\
     \midrule
     Ours & & \textbf{10.74} & \textbf{31.39} & & \textbf{11.26} & \textbf{31.73} & & \textbf{11.80} & \textbf{32.35} & & \textbf{12.14} & \textbf{32.86} & & \textbf{12.59} & \textbf{33.01} \\
     \bottomrule
    \end{tabular}
    }
    \caption{Quantitative evaluation of different methods for iterative local editing on our \textit{HieraFashion}. 
    TexFit-M refers to TexFit with our body part masks rather than its predicted ones. The best results are in bold.}
    \label{tab:editing}
\end{table*}

\begin{figure*}[!t]
  \centering
   \includegraphics[width=0.95\linewidth]{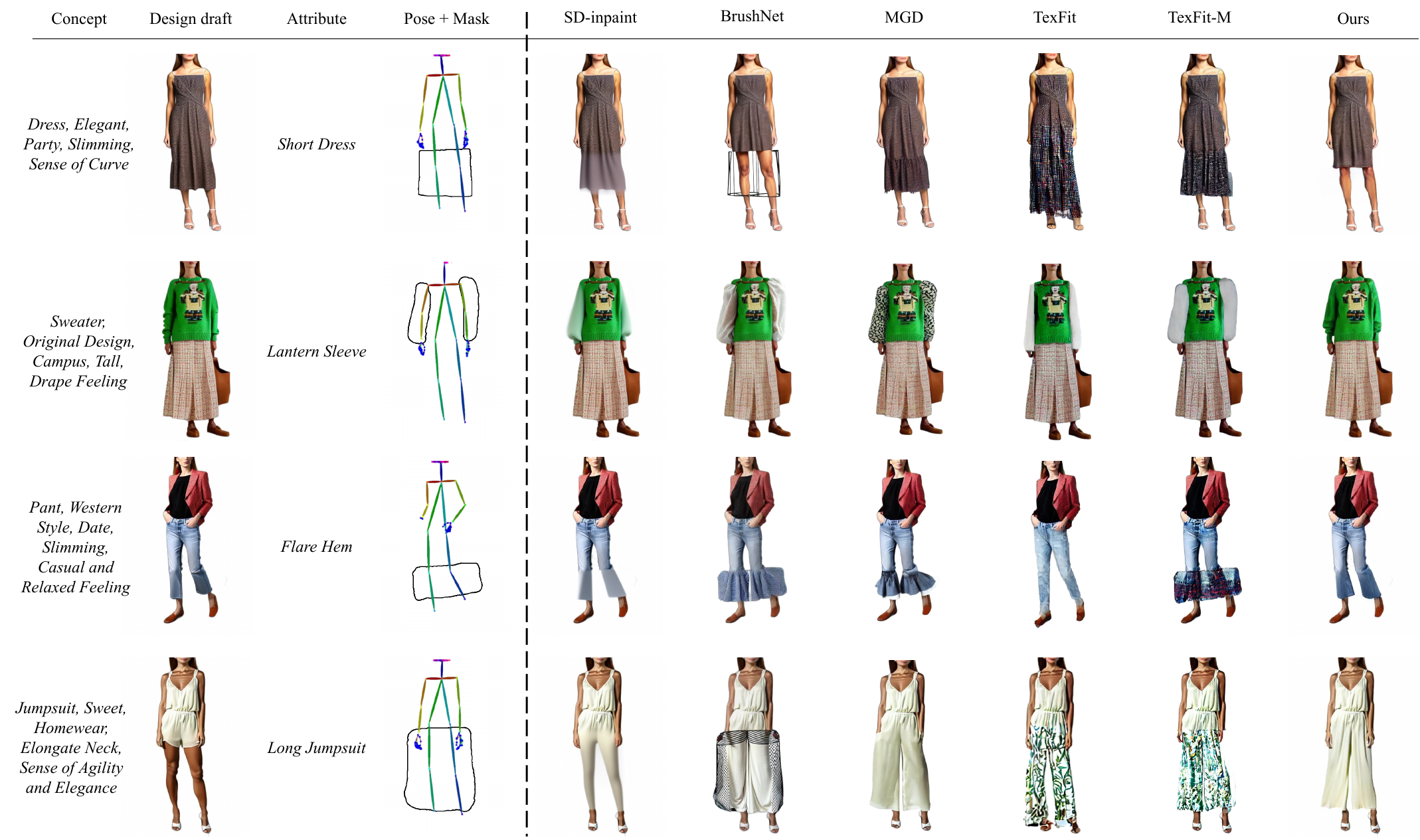}
   \caption{Qualitative comparison of iterative local editing.
    The latest editing methods often lack alignment with low-level attribute semantics, or cause undesirable global changes. Our method can precisely edit the corresponding regions according to the attribute descriptions while keeping the other regions unchanged, which is superior to other methods. 
    TexFit-M refers to TexFit with our body part masks rather than its predicted ones.}
   \label{fig:editing}
\end{figure*}

\noindent{\bf Evaluation metrics. }
To evaluate the realism and diversity of the generated images, we use the Fréchet Inception Distance (FID)~\cite{heusel2017gans} and Coverage (C)~\cite{naeem2020reliable} metrics.
For both metrics, we employ the CLIP ViT-B/32 model as the feature extractor.
Furthermore, to assess the coherence of the image to input prompts, we utilize the CLIP Score (CLIP-S)~\cite{HesselHFBC21}.
We fine-tune the CLIP ViT-B/32 model on the Dress Code Multimodal dataset and then on \textit{HieraFashoin} dataset to adapt to images and text descriptions in the fashion domain.
For fine-tuning the CLIP, we concatenate all the keywords for each image in our dataset into a single text description.

\subsection{Comparison to Prior Methods}
{\bf Quantitative Comparison. }
In Tab \ref{tab:generation}, we report the quantitative results of different methods on our \textit{HieraFashion} dataset for design draft generation given high-level prompts. 
As can be seen, our method consistently outperforms the competitors, in terms of realism and diversity (i.e, FID and Coverage) and prompt coherency (i.e, CLIP-S).

Tab \ref{tab:editing} shows the results of different methods for local editing. 
Our method outperforms the other methods across all the low-level apparel attributes. 
Moreover, as the number of editing iterations increases, the FID scores of all the  methods gradually decline.
This is because each editing iteration builds upon the previously generated image, resulting in gradually degraded visual quality.
Notably, our method is able to maintain consistently stronger performance through the iterations. 
In addition, TexFit-M using our body part masks is better than the original TexFit based on predicted editing regions, indicating that our approach of localizing target regions with body part masks is simple yet effective.

\begin{table*}
    \centering
        \scalebox{0.9}{
        \begin{tabular}{@{\hspace{2pt}}c@{\hspace{2pt}}c@{\hspace{2pt}}c@{\hspace{2pt}}c@{\hspace{2pt}}c@{\hspace{2pt}}c@{\hspace{2pt}}c@{\hspace{2pt}}c@{\hspace{2pt}}c@{\hspace{2pt}}c@{\hspace{2pt}}c@{\hspace{2pt}}c@{\hspace{2pt}}c@{\hspace{2pt}}c@{\hspace{2pt}}c@{\hspace{2pt}}c@{\hspace{2pt}}c@{\hspace{2pt}}c@{\hspace{2pt}}c@{\hspace{2pt}}}
        \toprule 
         \multirow{2}{*}{} & \multicolumn{3}{c}{\textbf{Draft Generation}} &  & \multicolumn{2}{c}{$\mathbf{A_1}$} &  & \multicolumn{2}{c}{$\mathbf{A_2}$} &  & \multicolumn{2}{c}{$\mathbf{A_3}$} &  & \multicolumn{2}{c}{$\mathbf{A_4}$} &  & \multicolumn{2}{c}{$\mathbf{A_5}$} \\
        \cmidrule{2-4} \cmidrule{6-7} \cmidrule{9-10} \cmidrule{12-13} \cmidrule{15-16} \cmidrule{18-19} 
        & \textbf{FID} $\displaystyle \downarrow $ &  \textbf{CLIP-S} $\displaystyle \uparrow $ & \textbf{C} $\displaystyle \uparrow $ &  & \textbf{FID} $\displaystyle \downarrow $ &  \textbf{CLIP-S} $\displaystyle \uparrow $ &  & \textbf{FID} $\displaystyle \downarrow $ &  \textbf{CLIP-S} $\displaystyle \uparrow $ &  & \textbf{FID} $\displaystyle \downarrow $ &  \textbf{CLIP-S} $\displaystyle \uparrow $ &  & \textbf{FID} $\displaystyle \downarrow $ &  \textbf{CLIP-S} $\displaystyle \uparrow $ &  & \textbf{FID} $\displaystyle \downarrow $ &  \textbf{CLIP-S} $\displaystyle \uparrow $  \\
        \midrule
        Flat & 14.29 & 28.17 & 0.43 &  & 16.79 & 29.21 &  & 18.13 & 30.09 &  & 19.39 & 30.36 &  & 20.48 & 30.55 &  & 21.34 & 30.83 \\
        RandAttrOrder & 13.81 & 28.98 & 0.59 &  & 14.26 & 29.46 &  & 14.87 & 30.11 &  & 15.52 & 30.61 &  & 16.37 & 30.80 &  & 16.95 & 30.96 \\
        \midrule
        Ours & \textbf{10.27} & \textbf{30.61} & \textbf{0.76} &  & \textbf{10.74} & \textbf{31.39} &  & \textbf{11.26} & \textbf{31.73} &  & \textbf{11.80} & \textbf{32.35} &  & \textbf{12.14} & \textbf{32.86} &  & \textbf{12.59} & \textbf{33.01} \\
         \bottomrule
        \end{tabular}
        }
        \caption{Effect of hierarchical descriptions and attribute ordering. We compare with two variants of our method (Ours): one conditioning on flat descriptions that combine high-level concepts and low-level attributes (Flat), and one using random order of attributes (RandomAttrOrder). The best results are in bold.}
        \label{tab:verify_hiera}
\end{table*}

\begin{figure}
  \centering
   \includegraphics[width=1.0\linewidth]{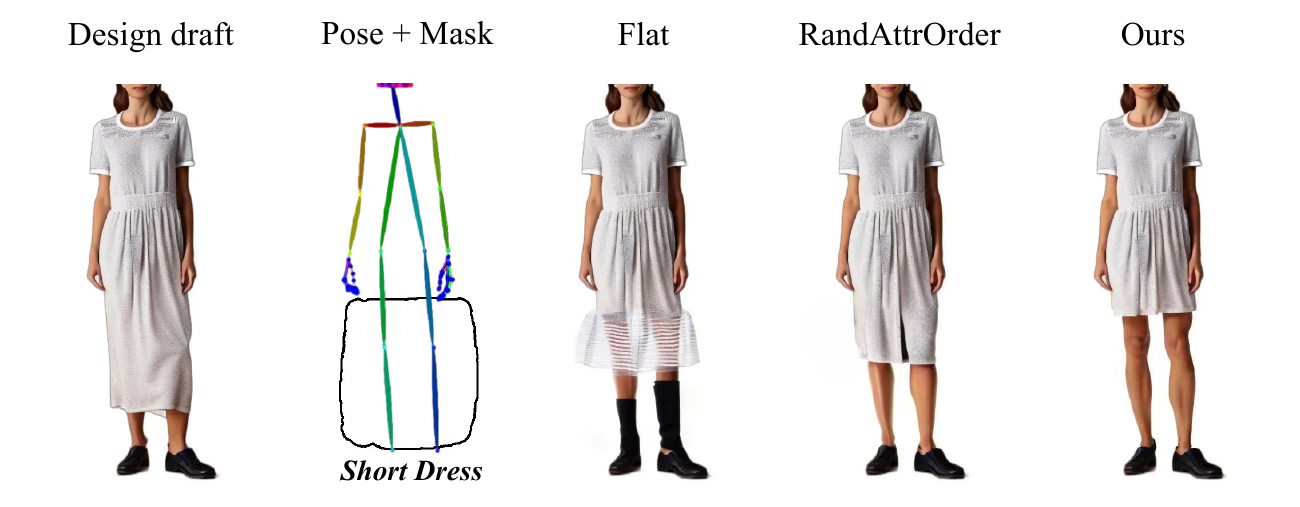}
   \caption{Comparison of our hierarchical model (Ours) against its flat (Flat) and random attribute ordering (RandAttrOrder) variants for local editing (long dress $\rightarrow$ short dress).}
   \label{fig:hiera_it}
\end{figure}

\noindent{\bf Qualitative comparison. }
In Fig \ref{fig:generation}, we provide a visual comparison of different methods in fashion draft generation. 
Our method can generate the designs that more faithfully convey the given concepts than the existing methods. 
Across different apparel categories, our method is able to synthesize designs with rich texture patterns and diverse apparel components. 
In contrast, CogView can synthesize simple apparels, but the generated results are blurry with many details lost.
The results by the SD are sharp with sufficient textural details, but are not matching the given concepts properly.
Attend-and-Exact improves upon CogView and the SD in terms of prompt alignment. 
However, there is still a noticeable gap between what its results convey and the given concepts. 
For example, the generated dress in the first row does not feel ``elegant" and the result in the second row does not ``elongate neck".

Fig \ref{fig:editing} compares the editing results of different methods. 
We observe two major shortcomings of the compared methods: first, they may struggle with producing the edits that precisely reflect the given attributes (e.g, the dress length is not reduced properly for all the other methods except BrushNet in the first row of Fig \ref{fig:editing}); second, they may generate the edited areas that are not harmonious with the rest part of the input design (e.g, the texture and color of the sleeve are changed in the second row of Fig \ref{fig:editing}).
In contrast, our method does not suffer from these issues, producing harmonious results that can accurately reflect different input attributes.

\subsection{Hierarchical Prompts and Attribute Ordering}
Our model considers hierarchical text descriptions with a multi-stage framework. 
We evaluate our hierarchical model against a flat alternative that collapses high-level concepts and low-level attributes into a single description, and conditions on the same description throughout the denoising process (Flat). 
We also try using random ordering of attributes in the iteration stage (RandAttrOrder), rather than our proposed ordering based on the size of influenced areas.

The results on both generation and editing tasks are shown in Tab \ref{tab:verify_hiera}.
Our full model is superior to the two alternatives across all the metrics for both tasks. 
The Flat performs the worst, giving very high FID scores and low CLIP-S scores, which confirms the importance of our hierarchical descriptions and multi-stage framework. 
Further, using random ordering of attributes also lead to inferior performance, suggesting the necessity of our proposed ordering strategy.
Fig \ref{fig:hiera_it} shows a visual comparison of different methods in an editing scenario to quantitatively demonstrate the advantage of our full model over the two variants.

\begin{figure}
  \centering
   \includegraphics[width=1.0\linewidth]{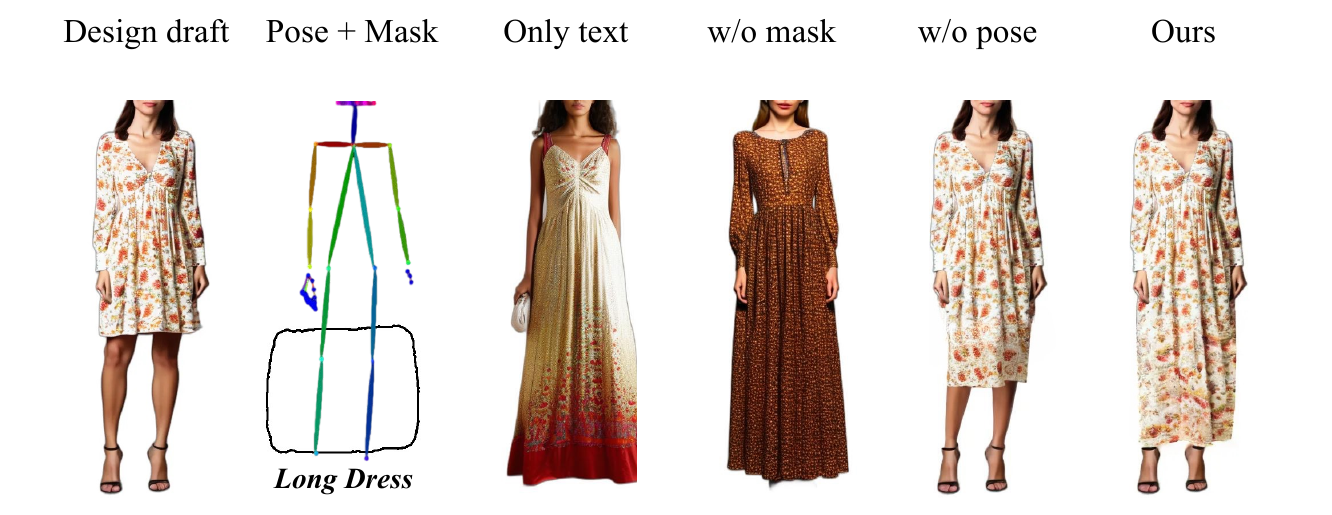}
   \caption{Effect of pose conditioning and body part masks for local editing (short dress $\rightarrow$ long dress).
   }
   \label{fig:ablation}
\end{figure}

\subsection{Ablation Study}

We further ablate two design choices for our local editing: 1) conditioning on pose maps; 2) using body part masks. 
Fig \ref{fig:ablation} provides a visual comparison, which shows the importance of these two components to visual quality. 
When the pose information is not used, the model fails to make the desired edit. 
Without the body part mask, the edited result involves large global modifications to the input image, totally changing the original style.
Removing one or both of the two components from our method lead to a degradation in both FID and CLIP-S across all the attributes (please see the supplemental for details).

\section{Conclusion}

In this paper, we propose a unified approach to facilitating fashion design based on a multi-stage diffusion model conditioned on hierarchical text descriptions. Our method naturally supports both design draft generation and iterative local editing, showing promising capability of fitting into and aiding in the typical fashion design workflow. 
We also contribute a fashion image dataset that comprise hierarchical text annotations, making it a valuable asset for training and evaluating various fashion design models. 
We hope that our idea of drawing analogy between the full fashion design pipeline and the denoising process of diffusion models, along with our dataset, can inspire and encourage future work in building practical systems for augmenting designers in the fashion domain and beyond.

\section{Acknowledgements}
This work is supported by the National Natural Science Foundation of China under Grant No. 62402306 and the Shanghai Natural Science Foundation of China under Grant No. 24ZR1422400.

\bibliography{aaai25}

\end{document}